\documentclass{article}

\usepackage[english]{babel}

\usepackage{amsmath}
\usepackage{amsfonts}
\usepackage{graphicx}
\usepackage{parskip}
\usepackage{comment}
\usepackage{float}
\usepackage{geometry}
\usepackage{xcolor}
\usepackage{hyperref}
\usepackage{subfig}
\usepackage{booktabs}
\usepackage{graphicx}
\newcommand{\signnn}{\hspace{-0.125cm}$^{***}$} 
 
\newcommand{\sign}{\hspace{-0.125cm}$^{*\ \ }$} 
\newcommand{\nosign}{\hspace{-0.125cm}$^{\ \ \ }$} 

\definecolor{ForestGreen}{RGB}{34,139,34}
\definecolor{MidnightBlue}{RGB}{25, 25, 112}

\hypersetup{
    colorlinks=true,
    urlcolor=MidnightBlue,
    linkcolor=ForestGreen,
    citecolor=ForestGreen
}

\usepackage[
backend=biber,
style= authoryear,
]{biblatex}

\bibliography{sample.bib}

 \newcommand\blfootnote[1]{%
  \begingroup
  \renewcommand\thefootnote{}\footnote{#1}%
  \addtocounter{footnote}{-1}%
  \endgroup
}

\usepackage{scalerel}
\usepackage{stackengine,wasysym}

\def\sym#1{\ifmmode^{#1}\else\(^{#1}\)\fi}

\title{Algorithmic progress in computer vision }
\author{
  Ege Erdil$^*$
  \and
  Tamay Besiroglu$^*$$^\dagger$
}
\date{%
    $^*$Epoch \hspace{0.15cm}
    $^\dagger$MIT FutureTech\\[2ex]%
}
\begin{document}
\maketitle
\begin{abstract}
We investigate algorithmic progress in image classification on ImageNet, perhaps the most well-known test bed for computer vision. We estimate a model, informed by work on neural scaling laws, and infer a decomposition of progress into the scaling of compute, data, and algorithms. Using Shapley values to attribute performance improvements, we find that algorithmic improvements have been roughly as important as the scaling of compute for progress computer vision. Our estimates indicate that algorithmic innovations mostly take the form of compute-augmenting algorithmic advances (which enable researchers to get better performance from less compute), not data-augmenting algorithmic advances. We find that compute-augmenting algorithmic advances are made at a pace more than twice as fast as the rate usually associated with Moore's law. In particular, we estimate that compute-augmenting innovations halve compute requirements every nine months (95\% confidence interval: 4 to 25 months).

\end{abstract}
\blfootnote{You can reproduce all our results using the code presented in \href{https://colab.research.google.com/drive/1-gBOhcVaYDNgO-9_EAOanfH7AeD17EGv?usp=sharing}{in this Colab notebook}.}
\blfootnote{We thank Matthew Barnett, Neil Thompson, Pablo Villalobos, Jaime Sevilla, Danny Hernandez, Marius Hobbhahn, Adam Papineau, Rick Korzekwa, and Lawrence Phillips for their helpful comments. We are also grateful to Owen Dudney for his assistance in building the relevant data-sets.}
\section{Introduction}
It is a matter of debate how much of the recent progress in machine learning has come from the scaling of compute and the sizes of models and data-sets, and how much has come from improvements in the underlying algorithms and architectures. In this work, we provide a decomposition of these different sources of progress on the task of image classification on the well-known ImageNet dataset, and provide insights into how algorithmic advances produce better models.\par

\begin{figure}[h]%
    \centering
    \subfloat[\centering Pareto frontiers in data and compute for AlexNet performance]{{\includegraphics[width=5.55cm]{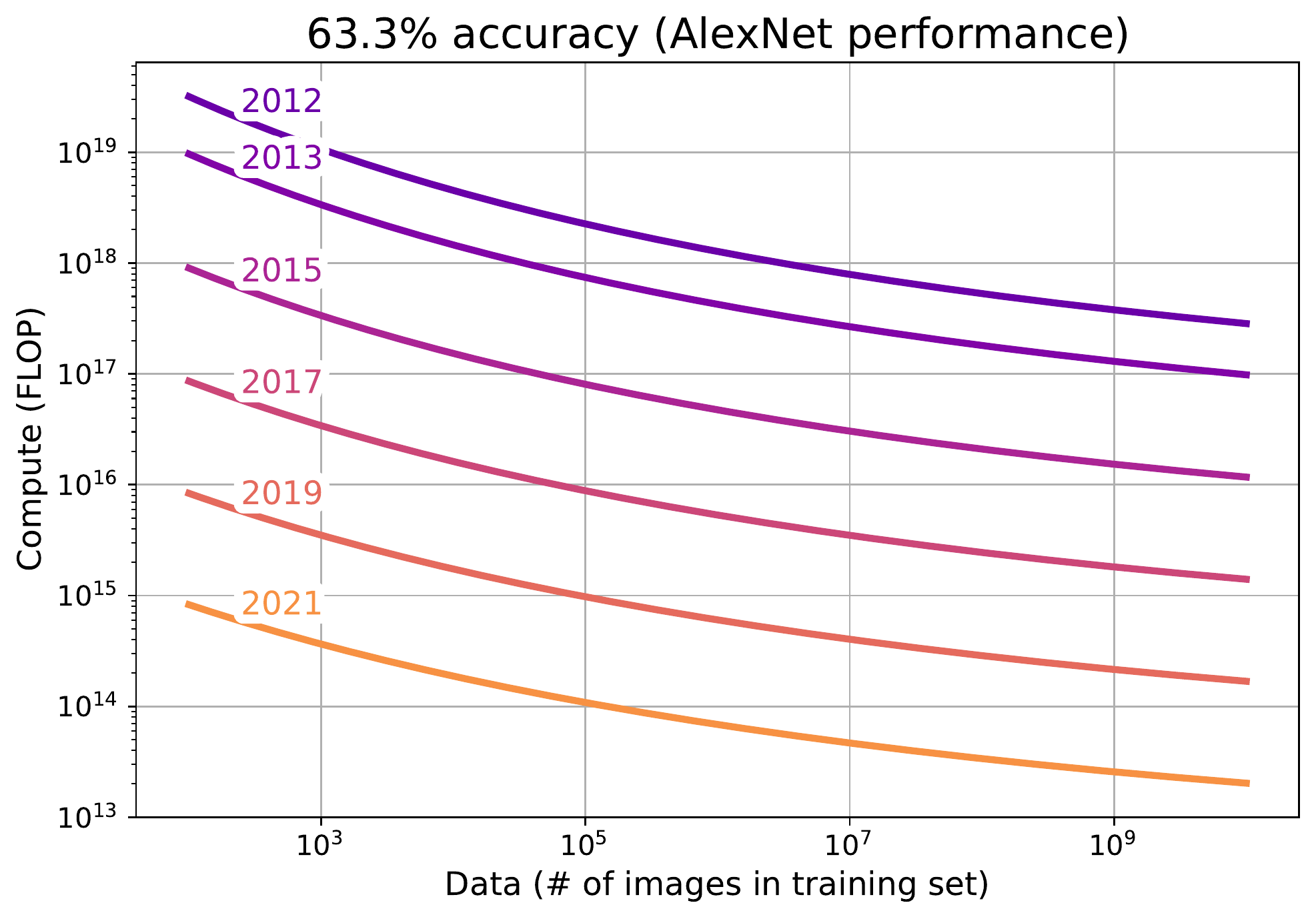} }}%
    \qquad \hspace{-0.75cm}
     \subfloat[\centering Pareto frontiers in data and compute for ResNeXt-101 performance]{{\includegraphics[width=5.55cm]{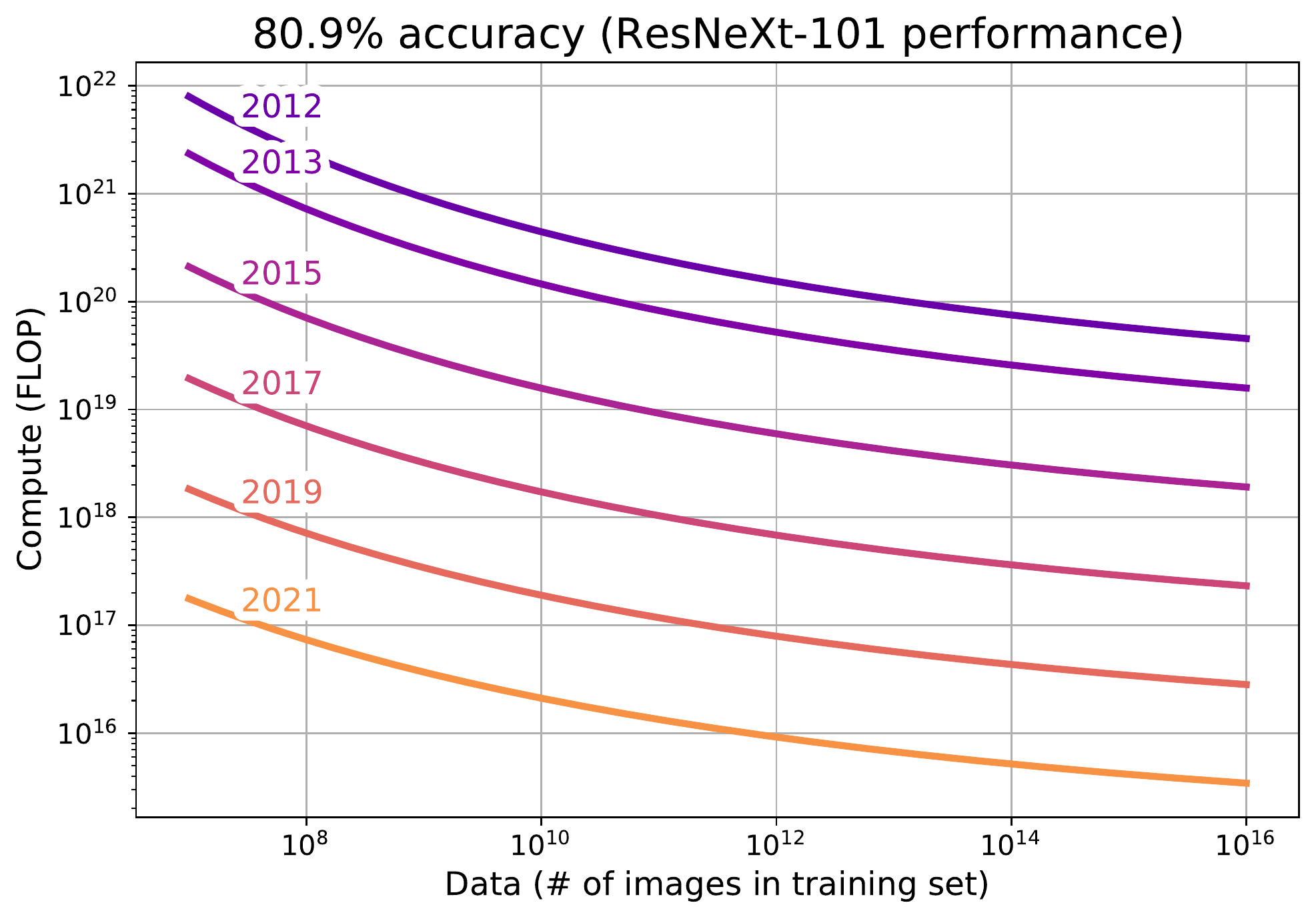}}}%
    \qquad \hspace{-0.75cm}
    \subfloat[\centering Pareto frontiers in data and compute for ViT-e performance]{{\includegraphics[width=5.55cm]{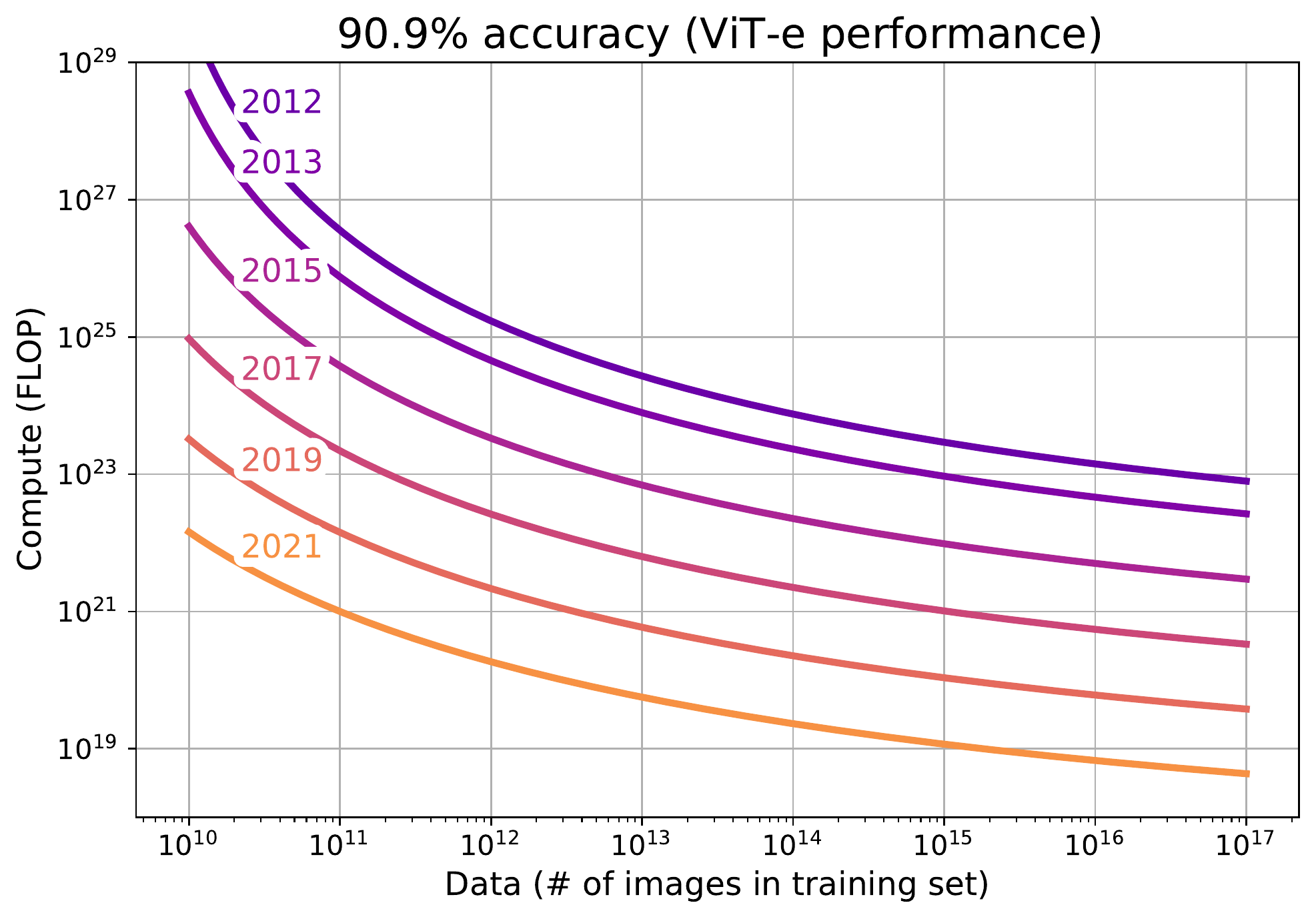}}}%
    \caption*{\small \centering \textbf{Figure 1. Pareto frontiers for training models to achieve performance of well-known models over time.} Our estimates indicate that compute-augmenting algorithmic improvements double the effective compute available to train image classification models every nine months. Note that these plots are predictions made by a model, and could be extrapolating the model beyond its domain of validity (see \ref{sec:model} for additional discussion of this caveat).}%
    \label{fig:example}%
\end{figure}

Attempts at providing a decomposition of sources of advances in machine learning predate our work. Recently, \cite{karpathy} investigates this question by taking the neural network from \cite{lecun1989backpropagation} and progressively augmenting it with modern techniques, such as dropout, the AdamW optimizer (instead of SGD) and ReLU (instead of tanh), among other things. The result is that without using additional compute or data, the author reports being able to reduce the test error rate of the classifier by more than a factor of two, which suggests that algorithmic improvements have certainly contributed to the success of modern computer vision. \cite{hernandez2020measuring} re-implement 15 open source popular models with few hyperparameter adjustments and find a 44-fold reduction in the compute required to reach the same level of performance as AlexNet (\cite{krizhevsky2012imagenet}). This suggests that algorithmic progress is slightly more important than hardware improvements, as the rate of algorithmic efficiency improvements outpaces the original Moore’s law rate of improvement in hardware efficiency.

Outside of machine learning, there are a larger number of relevant results on the relative importance of algorithmic and hardware improvements, much of which indicates that hardware and software advances have been roughly of similar importance across a wide range of computing problems. In their work, \cite{sherry2021fast} analyse data from 57 textbooks and more than 1137 research papers on algorithmic improvements and find a high degree of variation in the pace of algorithmic improvements. They find that while around half of algorithm families—algorithms for a given problem—experience little to no improvement, algorithm families that see improvements typically improve at a pace comparable to or faster than Moore's law, at least for moderately-sized problems. Similarly, \cite{grace2013algorithmic} investigates algorithmic progress in six domains (SAT solving, Game Playing, Factoring, Physics Simulations, Operations Research, and Machine Learning) and finds that, while the data is often noisy, gains from algorithmic progress have been roughly fifty to one hundred percent as large as those from hardware progress. This suggests that, while there is much variation, algorithmic improvements can be at least as important as hardware improvements for a wide range of problems.

\cite{koch2022progress} investigate the relative importance of hardware and algorithmic improvements in the progress computational methods for solving Linear Programs (LP) and Mixed Integer Linear Programs (MILP). By running new solvers on old hardware and old solvers on new hardware, they find that computer hardware got about 20 times faster, and the algorithms improved by a factor of about nine for LP and around 50 for MILP. Their work suggests that, although there is some variance, hardware and algorithmic progress were of the same order. Similarly, \cite{fichte2020time} evaluates the relative importance of algorithmic and hardware improvements for progress in SAT-solvers in the past 20 years, and finds that progress on the algorithmic side has at least as much impact as the progress on the hardware side. For computer chess, \cite{Adamczewski} investigates the improvement in chess engines while keeping clock-speed fixed, and finds that hardware has been roughly twice as important as software in explaining historical chess progress.

This article is organized as follows. In Section 2, we discuss the general problem of measuring the extent of algorithmic advances in computing and in machine learning in particular, and highlight key methodological issues that arise when trying to derive meaningful general results about the extent of algorithmic innovation. We then develop an empirical approach to modeling how performance improvements relate to the algorithmic improvements and the scaling of compute and data in Section 3. After describing our dataset in Section 4, we present our empirical analysis in Section 5. First, we provide a decomposition of performance improvements into algorithmic improvements, compute- and data-scaling for well-known computer vision models. We then provide a further decomposition wherein we distinguish between compute- and data-augmenting algorithmic progress, and show that most algorithmic progress is of the compute-augmenting type. Finally, we estimate the rate of algorithmic innovation by inferring how much less compute is required over time to train models that achieve a certain level of performance. 

\section{Estimating the extent of algorithmic advances}
\label{sec:extent}

In contrast to other more traditional algorithm families, machine learning models typically resists analysis of algorithmic complexity. That is, we generally cannot definitely say how many training steps a model needs to perform in order to achieve a given level of performance on a dataset. Nevertheless, there are a few different ways to estimate algorithmic progress for machine learning. One idea, used in \cite{hernandez2020measuring}, is to fix a specific threshold (e.g., accuracy or loss) in a benchmark that's been attained by an earlier model and then look at how much compute newer architectures need to achieve the same level of performance on the same dataset. The compute savings then give us a way to quantify the improvement: an algorithm or architecture is two times better than another one if it achieves the same threshold in the benchmark with half the compute. Fixing the threshold to be the level of performance achieved by \cite{krizhevsky2012imagenet} in top-5 ImageNet classification error, they find that the level of compute required to achieve this performance halves every 16 months.

We could naively use this as a measure of the growth in algorithmic efficiency by saying that it improves at a rate of $ 2^{12/16} - 1 \approx 68.2\% $ per year. However, it is unclear how valid this interpretation is. The problem is that this definition of growth is highly sensitive to the exact benchmark and threshold pair that we choose. Choosing easier-to-achieve thresholds makes algorithmic improvements look less significant, as the scaling of compute easily brings early models within reach of such a threshold. By contrast, selecting harder-to-achieve thresholds makes it so that algorithmic improvements explain almost all of the performance gain. This is because early models might need arbitrary amounts of compute to achieve the performance of today's state-of-the-art models. Concretely, in \hyperref[sec:Sensitivity]{Appendix C}, we show that the estimates of the pace of algorithmic progress from \cite{hernandez2020measuring} might vary by around a factor of ten, depending on whether an easy or difficult threshold is chosen.

In addition, we also have to take into account the concern brought up in \cite{paulfchristiano} that some algorithmic improvements are \textit{absolute} in the sense that they are better regardless of the scale at which the algorithm is executed (such as using quicksort instead of bubble sort), while some are about adapting to new hardware. For example, \cite{karpathy} finds that using a miniature Vision Transformer at compute parity with the convolutional neural net from \cite{lecun1989backpropagation} fails to achieve good performance. So it could be the case that the good performance of modern computer vision models is partly explained by new Transformer-based architectures, even though such architectures might need \textit{more} compute to achieve the same level of performance as older models.

To give an analogy from algorithmic complexity theory, a $ \Theta(N) $ time complexity algorithm with a large constant can be worse than a $ \Theta(N^2) $ time complexity algorithm with less pre-processing unless $ N $ is large enough, and if we do not have enough compute to use either algorithm when $ N $ is large we might simply go with the asymptotically less efficient algorithm. It seems plausible that this is relevant to actual algorithmic innovation within machine learning. As we scale up models and training data, it becomes more and more important to use algorithms which have favorable scaling behaviour even if their low-parameter, low-data performance is inferior to previous architectures. This favorable scaling behavior might manifest both in forms of overall compute efficiency, but also in the form of relying heavily on optimized matrix multiplications and other parallelizable pieces in new models, as parallelizing the same amount of training compute becomes an important source of algorithmic efficiency. 

These consideration suggest that we should consider models implemented at scales that bring out these various advantages (such as favourable scaling behaviour, or conversely, their favourable low-parameter, low data-performance). One place to look for this is the actual implementations as presented in the relevant publications, which is the approach pursued in this work.

\section{Empirical approach}
\label{sec:empirical}

In this section, we describe the empirical approach, informed by neural scaling laws, for estimating how compute, data, and the quality of algorithms and architectures effect the performance of computer vision models. In what follows, we describe the empirical model that we use to produce our main results. 

\subsection{Empirical model}
\label{sec:model}

In developing an empirical model to estimate the relation of model-performance to data, compute and algorithms, we have the following desiderata. Firstly, the specification of the model should be consistent with prior work on the relation between compute, data, and performance, particularly the literature on neural scaling laws (such as \cite{hoffmann2022training}). Secondly, the model should enable us to infer the importance of different mechanisms through which algorithmic innovations enable performance improvements, namely, whether algorithmic improvements work primarily through relaxing data bottlenecks or compute bottlenecks. Lastly, standard goodness-of-fit considerations are taken into account.

The model we use to get the main conclusions of this paper is as follows:
\begin{equation}
   \sigma^{-1}(P) = \sigma^{-1}(\sigma(C) \times \sigma(D)) + \varepsilon,\hspace{0.15cm} \text{where} \hspace{0.15cm} \varepsilon \sim N(0, \delta^2). 
\end{equation}
Here, \( \sigma: x \to 1/(1 + e^{-x}) \) is the logistic function, \( \sigma^{-1}: p \to \log(p/(1-p)) \) is its inverse function, and \( P \in (0, 1) \) stands for our metric of performance, which in this case is top-1 accuracy on the ImageNet test-set. $C$ denotes the effective compute budget and $D$ denotes the effective data budget in the model (both in logarithms), which are defined as follows:
\begin{align}
C &= \alpha_1 + \alpha_{\textrm{Year}} \times (\textrm{Year} - 2012) + \alpha_{\textrm{compute}} \times \log \left(\textrm{compute}\right),\\
D &= \beta_1 + \beta_{\textrm{Year}} \times (\textrm{Year} - 2012) + \beta_{\textrm{data}} \times \log \left( \text{data}\right) .
\end{align}

If we imagine setting the noise term to zero, the model implies that \( P \) behaves as the product of two sigmoid terms. The model can be shown to approximate to a familiar functional form that others have found in prior work on neural scaling laws, i.e., compare the following approximations implied by our model to those found by \cite{hoffmann2022training}:
\begin{equation}
    \underbrace{1-P \approx \frac{\tilde{A}}{C^{\alpha_{\text{compute}}}} + \frac{\tilde{B}}{D^{\beta_\text{data}}}}_{\text{Our Model}}, \hspace{0.35cm}      \underbrace{L \approx \frac{A}{N^{\alpha}} + \frac{B}{D^{\beta}}}_{\text{\cite{hoffmann2022training}}},
\end{equation}
where $\tilde{A} \equiv \exp\big(-\alpha_1 - \alpha_{\textrm{Year}}\times (\textrm{Year} - 2012)\big)$, and $\tilde{B} \equiv \exp\big(-\beta_1 -\beta_{\textrm{Year}}\times (\textrm{Year} - 2012)\big)$. Note that these are of a similar form to the results from \cite{hoffmann2022training}, with two main differences. Firstly, while \cite{hoffmann2022training} has a model-size term, we instead have a compute term, as our model achieves a better fit with this substitution. Secondly, in our model, $A$ and $B$ terms are not constant but vary over time, to capture the notion of algorithmic progress improving how effectively compute and data is used to train an effective model.

Throughout, compute refers to the number of FLOP performed, and data refers to the number of images in the training set used to train the relevant model. We normalize compute and data by dividing these by \(4.7 \times 10^{17} \) and \( 1.28 \times 10^6 \) respectively, as these are the values of compute and data used in \cite{krizhevsky2012imagenet}. This is done in order for AlexNet to be a natural baseline, as well as to simplify the associated maximum likelihood estimation problem. 

This model might not extrapolate well when compute \( C \) and data \( D \) are taken to be far from the range we observe them to be in whatever dataset we've fit the model to. Consequently, care should be taken when using the model to make predictions that might be beyond its domain of validity. As one example, this model can make predictions about the performance of models with \( D \gg C \), which is meaningless as it's not possible to process some number of images without at least doing on the order of that many floating point operations.

In estimating our model, we enforce the following prior distributions for each parameter. These distributions have zero mean, and their variance is chosen on the basis that these achieve the highest mean log scores in one-out cross validation:
\begin{equation}
    \alpha_{1}, \beta_{1}, \log(\sigma)  \sim \text{Normal}(0,1), \hspace{0.15cm}  \alpha_{year}, \alpha_{\text{compute}}, \beta_{year}, \beta_{data} \sim \text{Normal}(0,0.09).
\end{equation}
Note that the prior distributions for the parameters that enter as intercepts ($\alpha_1, \beta_1$) have lower variances than those for parameters that enter multiplicatively. We then use maximum a posteriori estimation, which is equivalent to doing maximum likelihood estimation with \( L^2 \) regularization penalties added to the objective. The reason for this choice is that the relatively small size of our dataset means the parameters about data scaling end up being poorly identified in the absence of any regularization; and when we bootstrap to get standard errors on the parameters with pure MLE, the data scaling parameters are set to unrealistic values for \( \sim 5 \% \) of simulated datasets generated during the bootstrap.

In order to be able to estimate this model properly, it is therefore important to somehow incorporate a prior into the estimation. The alternative would have been to use a smaller model where the parameters do end up being identified better, but it is hard to construct a realistic model which has algorithmic progress acting on both data and compute bottlenecks without at least having this much complexity. Still, to avoid problems with overfitting, we chose the priors to have relatively high variance. This was enough to eliminate the undesirable behavior with multiple local minima. 

\section{Dataset}
\label{sec:data}

We extend the dataset from \cite{thompson2020computational} consisting of 124 computer vision models tested on the ImageNet-1k data set (\cite{deng2009imagenet}). Since most papers do not report the total amount of compute they used during their training run, compute estimates are instead based on the model architecture, dataset size and the number of epochs the model has been trained for. The compute estimates are derived from the underlying papers following the procedure described by \cite{sevillacompute}, which is summarized in \hyperref[sec:estimatingcompute]{Appendix A}.\footnote{The dataset we use may be found on \href{https://github.com/Besiroglu/computer_vision_data/blob/main/imagenet.csv}{this GitHub repository}.}

If multiple models are presented in a single paper, we include at most three of the top models from each paper in our dataset. We exclude re-implementations of prior models from our dataset because we find the reported performance to be unreliable, and because these might benefit from software-related improvements such as high-level libraries, interpreters and compilers as well as improvements to frameworks for widely used deep neural networks such as cudNN and cublas, which mean that re-implementations might unknowingly benefit from software improvements.\footnote{We thank Marius Hobbhahn for pointing this out to us.} We further exclude models that involve neural architecture search, as these models are likely to have much different scaling behavior than other types of models.
 
\section{Results}
\label{sec:results}

In this section, we first present the main parameter estimates of the mixture model in several pieces. We then move on to several sections in which we'll discuss various implications of the model. 

\subsection{Empirical estimates}
Recall that the model that we developed in our Empirical Approach section is:
\begin{equation}
   \sigma^{-1}(P) = \sigma^{-1}(\sigma(C) \times \sigma(D)) + \varepsilon,\hspace{0.15cm} \text{where} \hspace{0.15cm} \varepsilon \sim N(0, \delta^2),
\end{equation}
where the effective compute- and data-budgets, $C$ and $D$ are defined as follows:
\begin{align}
C &= \alpha_1 + \alpha_{\textrm{Year}} \times (\textrm{Year} - 2012) + \alpha_{\textrm{compute}} \times \log \left(\textrm{compute}\right),\\
D &= \beta_1 + \beta_{\textrm{Year}} \times (\textrm{Year} - 2012) + \beta_{\textrm{data}} \times \log \left( \text{data}\right). 
\end{align}

Our estimates are summarized in Table 1. We find that $\alpha_{\textrm{Year}}$ is positive and statistically significant at the 5\% level, indicating that, over time, better models use compute more effectively. By contrast, $\beta_{\textrm{Year}}$ is much smaller and not statistically significant, which indicates that the algorithmic improvements that occur over time have no discernible net effect on expanding effective data budgets. 
\begin{figure}[!htb]
    \centering
    \begin{minipage}{.5\textwidth}
        \centering
        \begin{tabular}{@{}lcc@{}}
\toprule
 & Estimate & 95\% CI \\ \midrule
$\alpha_1$ & \begin{tabular}[c]{@{}c@{}}$\underset{(0.313)}{0.889}$ \signnn \end{tabular} & $0.584, 1.920$ \\
$\alpha_{\textrm{Year}}$ & \begin{tabular}[c]{@{}c@{}}$\underset{(0.045)}{0.159}$ \sign\\ \end{tabular} & $0.028, 0.212$ \\
$\alpha_{\textrm{compute}}$ & \begin{tabular}[c]{@{}c@{}}$\underset{(0.045)}{0.154}$ \signnn\\\end{tabular} & $0.075, 0.229$ \\
$\beta_1$& \begin{tabular}[c]{@{}c@{}}$\underset{(0.285)}{1.783}$ \signnn\\ \end{tabular} & $1.194, 2.315$ \\
$\beta_{\textrm{Year}}$ & \begin{tabular}[c]{@{}c@{}}$\underset{(0.034)}{0.019}$ \nosign\\ \end{tabular} & $-0.036, 0.089$\\
$\beta_{\textrm{data}}$ & \begin{tabular}[c]{@{}c@{}}$\underset{(0.0140)}{0.063}$ \signnn\\ \end{tabular} & $0.038, 0.089$ \\
$\log(\delta)$ & \multicolumn{1}{l}{\begin{tabular}[c]{@{}l@{}}$\underset{(0.080)}{-2.160}$\nosign\\ \end{tabular}} & \multicolumn{1}{l}{$-2.410, -2.090$} \\
\begin{tabular}[c]{@{}l@{}}Log \\ likelihood\end{tabular} & 267.062 & — \\ \bottomrule
\end{tabular}
\caption*{\small \centering  \textbf{Table 1. Parameter estimates}}
        \label{fig:prob1_6_2}
    \end{minipage}%
    \begin{minipage}{0.5\textwidth}
        \centering
        \includegraphics[scale=0.61]{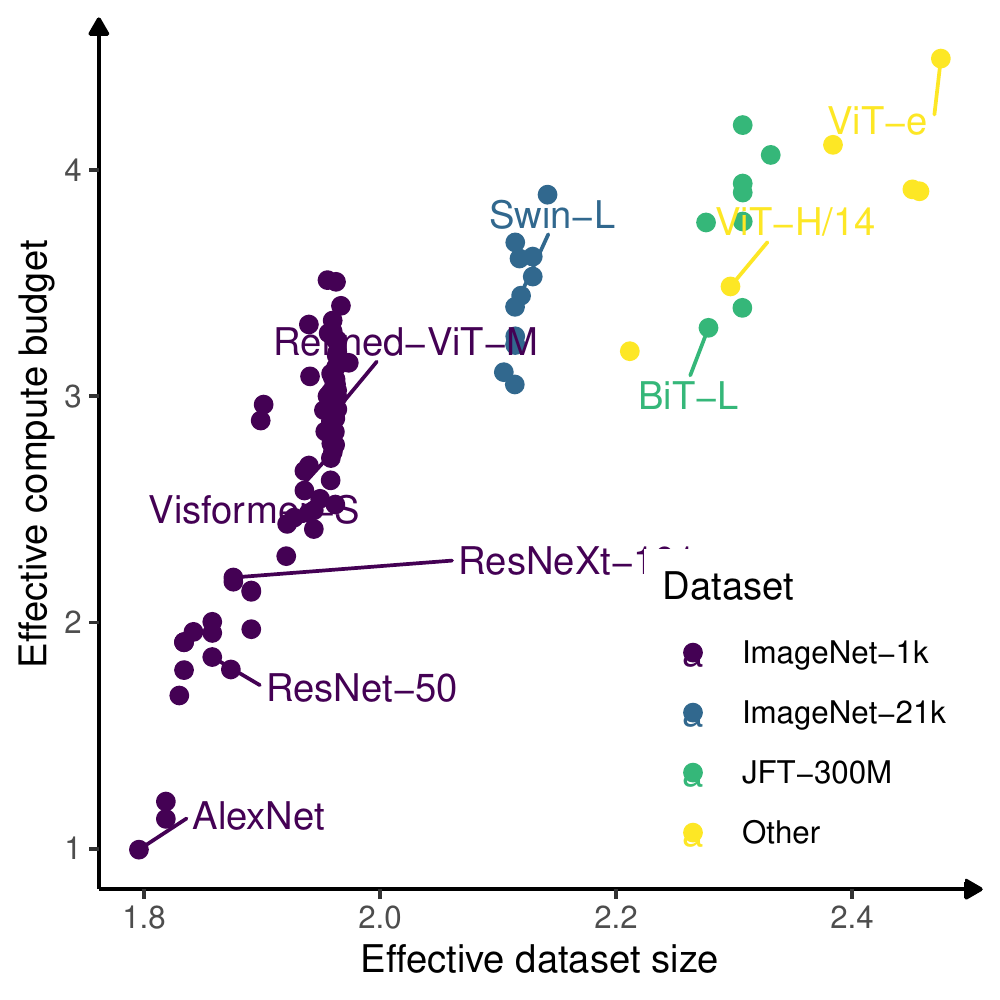}
        \caption*{\centering \small \textbf{Figure 2. Predicted effective dataset and effective compute budgets}}
        \label{fig:prob1_6_1}
    \end{minipage}
\vspace{0.15cm}

\small{\textbf{Table 1 and Figure 1.} The confidence intervals in Table 1, which are obtained by dividing the point-estimate by standard errors, are unreliable as we find that the parameter distributions are not uni-modal and highly skewed. We therefore also report \(90\% \) confidence intervals for all of the parameter estimates by bootstrapping 100 iterations. *, **, and ** correspond to p values below 5\%, 1\% and 0.01\% respectively. Figure 1 indicates that the growth in effective compute was around 20-fold greater than the growth in effective data budgets when we compare the smallest and largest models in our dataset.\footnotemark}
\end{figure}
\footnotetext{That is, effective compute budgets expanded roughly 4.5-fold (in log-space), while data budgets expanded roughly 1.4-fold (also in log-space). Hence, effective compute outgrew effective data by a factor of $\exp(4.5)/\exp(1.4) \approx 20$.}
\subsection{Progress has mostly been due to compute and algorithms}

Given that our model is nonlinear, simple attributions of performance improvements to the scaling of compute, data, and improvements in algorithms, based on coefficient ratios is not possible. Hence, we use Shapley values (\cite{hart1989shapley})—the average expected marginal contributions—to compute the relative importance of each of these contributing factors.

The key idea is this: let $f$ be an accuracy model, i.e., a function that maps its arguments to the point estimate of top-1 accuracy that would be attained by the model having those properties. Performance attribution is difficult because if we imagine going from $f(\textrm{Year}_1, \textrm{Data}_1, \textrm{Compute}_1)$ to $ f(\textrm{Year}_2, \textrm{Data}_2, \textrm{Compute}_2)$ by substituting out one variable at a time and measuring the improvement we get at each step, the fraction of the overall improvement we will attribute to a particular variable depends on the order in which we do the substitutions. It has been shown that this natural averaging solution to the problem has many other desirable properties (such as efficiency, symmetry, and linearity amongst others), and that it is the only way of attributing performance that meets some of these properties at once (\cite{hart1989shapley}).\footnote{It is a standard idea in representation theory that if an object is not invariant under the action of a group \( G \), we can often make it invariant by averaging its value over its orbit under the action of \( G \) (\cite{neusel2002invariant}). In our context \( G = S_3 \) is the symmetric group on three letters consisting of the six different ways in which three letters may be permuted, accounting for the six different orders in which we might try to substitute the new variables for the old. Thus, we can simply compute the improvement we would attribute to any one of them for a specific permutation $\sigma \in S_3$ and then average over all such $\sigma$ to get our final answer.}

\begin{table}[h!]
    \centering
\begin{tabular}{@{}lcccc@{}}
\toprule
 & \multicolumn{1}{l}{\begin{tabular}[c]{@{}l@{}}Reduction\\ in error\end{tabular}} & \multicolumn{1}{l}{\begin{tabular}[c]{@{}l@{}}Algorithmic \\ progress\end{tabular}} & \multicolumn{1}{l}{\begin{tabular}[c]{@{}l@{}}Compute \\ scaling\end{tabular}} & \multicolumn{1}{l}{\begin{tabular}[c]{@{}l@{}}Data \\ scaling\end{tabular}} \\ \midrule
AlexNet $\rightarrow$ ResNet50 & 23.7 & 64.9\% & 35.1\% & — \\
AlexNet $\rightarrow$ ResNeXt-101 & 24.0 & 70.6\% & 29.3\% & — \\
AlexNet $\rightarrow$ BiT-L & 24.2 & 40.8\% & 47.2\% & 12.1\% \\
AlexNet $\rightarrow$ VIT-H/14 & 24.8 & 43.7\% & 44.4\% & 11.9\% \\
AlexNet $\rightarrow$ VIT-e & 27.6  & 41.6\% & 43.6\% & 14.8\% \\
ResNet50 $\rightarrow$ BiT-L & 10.4 & 30.7\% & 47.3\% & 22.0\% \\
ResNet50 $\rightarrow$ VIT-H/14 & 10.9 & 35.2\% & 43.4\% & 21.4\% \\
ResNet50 $\rightarrow$ VIT-e & 13.8 & 34.1\% & 40.9\% & 25.0\% \\
ResNeXt-101 $\rightarrow$ BiT-L & 6.6 & 24.9\% & 49.8\% & 25.4\%\\
ResNeXt-101 $\rightarrow$ VIT-H/14 & 7.2 & 30.1\% & 45.3\% & 24.5\% \\ 
ResNeXt-101 $\rightarrow$ VIT-e & 10.0 & 30.3\% & 41.6\% & 28.1\% \\ \bottomrule
\end{tabular}
    \caption*{\centering \small \textbf{Table 2.} Attribution of progress to algorithmic progress, compute scaling and data scaling between early well-known models and newer models based on Shapley decomposition. ``NS" indicates that there was no scaling of the relevant input between these models. Numbers may not all add up to 100 due to rounding.}
    \label{tab:my_label}
\end{table}

We find that in most cases, improvements in performance have been between 25\% and 70\% the result of algorithmic innovation: the lowest contribution of algorithmic innovation we predict is between ResNeXt-101 to BiT-L at 24.9\%, while the highest contribution is predicted to be made between AlexNet and ResNeXt-101 at 70.6\%. The estimates of the fraction of improvements due to the scaling of compute is fairly tightly clustered in the 30\% to 55\% range. We find our results are robust to using Shapley values in log odds space (i.e., after applying an inverse sigmoid transformation), see \hyperref[sec:supporting]{Appendix F}.

These estimates indicate algorithmic improvements in image classification have been roughly as important as the scaling of compute for driving performance improvements (with the caveat that algorithmic progress was more crucial in early years, and compute scaling more important in later years). This result is consistent with work in other domains, such as computational solvers for linear programs (\cite{koch2022progress}, SAT-solvers \cite{fichte2020time}), as well as computer chess \cite{Adamczewski}, for which it is found that software and hardware improvements are both important drivers of performance improvements, and that hardware improvements generally account for slightly more of total performance improvements.

\subsection{Most algorithmic progress involves improved compute efficiency}
\label{sec:compute_algo}

Our estimates, summarized in Table 1, indicate that, on the margin, algorithmic innovation expands the effective compute budget an equivalent of $\hat{\alpha}_{\text{Year}}/\hat{\alpha}_{\text{Compute}} \approx 100.96\%$ each year [95\% CI: 24.60\%, 215.18\%], while expanding the effective data budget by $\hat{\beta}_{\text{Year}}/\hat{\beta}_{\text{data}} \approx 37.64\%$ per year [95\% CI: -46.95\%, 134.46\%]. Hence, our estimates suggest that the main effect of algorithmic improvements is the expansion of effective compute budgets, i.e., algorithmic progress is primarily compute-augmenting.

However, given that the model is non-linear, small expansions of the effective data budget could have relatively large effects on model-performance. Hence, while our parameter estimates suggest algorithmic innovations are more effective at augmenting compute budgets relative to data budgets, these estimates alone do not necessarily imply this is also the primary mechanism through which algorithmic innovation improve performance.

Looking at performance attributions reveals further evidence about the types of algorithmic improvements that are most important for explaining variation in model performance. To do so, we once again rely on the Shapley value concept to estimate the average expected marginal contributions of compute-augmenting and data-augmenting algorithmic innovations.

\begin{table}[h!]
    \centering
\begin{tabular}{@{}lcccc@{}}
\toprule
 & \multicolumn{1}{l}{\begin{tabular}[c]{@{}l@{}}Reduction\\ in error\end{tabular}} & \multicolumn{1}{l}{\begin{tabular}[c]{@{}l@{}}Algorithmic progress\\(data augmenting)\end{tabular}} & \multicolumn{1}{l}{\begin{tabular}[c]{@{}l@{}}Algorithmic progress\\(compute augmenting)\end{tabular}} & \multicolumn{1}{l}{\begin{tabular}[c]{@{}l@{}}\end{tabular}} \\ \midrule
AlexNet $\rightarrow$ ResNet50 & 23.7 & 5.2\% & 59.7\%\\
AlexNet $\rightarrow$ ResNeXt-101 & 24.0 & 5.9\% & 64.7\% \\
AlexNet $\rightarrow$ BiT-L & 24.2 & 4.9\% & 35.9\% \\
AlexNet $\rightarrow$ VIT-H/14 & 24.8 & 5.4\% & 38.3\%  \\
AlexNet $\rightarrow$ VIT-e & 27.6  & 5.6\% & 36.0\% \\
ResNet50 $\rightarrow$ BiT-L & 10.4 & 5.1\% & 25.6\% \\
ResNet50 $\rightarrow$ VIT-H/14 & 10.9 & 6.0\% & 29.2\% \\
ResNet50 $\rightarrow$ VIT-e & 13.8 & 6.4\% & 27.7\% \\
ResNeXt-101 $\rightarrow$ BiT-L & 6.6 & 4.4\% & 20.5\% \\
ResNeXt-101 $\rightarrow$ VIT-H/14 & 7.2 & 5.5\% & 24.6\% \\ 
ResNeXt-101 $\rightarrow$ VIT-e & 10.0 & 6.1\% & 24.2\%  \\ \bottomrule
\end{tabular}
    \caption*{\centering \small \textbf{Table 3.} Shares of algorithmic progress that is compute- vs. data-augmenting based on Shapley value decomposition. The sum of the relevant fractions equals the total share of performance improvement due to algorithmic progress, and therefore adds up to <100\%.}
    \label{tab:my_label}
\end{table}

Our estimates indicate that the overwhelming majority of expected marginal contributions from algorithms occurs through augmenting compute, rather than through augmenting the use of data. Specifically, our model indicates that compute-augmenting algorithmic innovations explain between four and 11-fold more variation in performance than do data-augmenting algorithmic innovations.

Interestingly, we find that the relative compute-augmenting algorithmic innovations have perhaps declined (if we compare, e.g., AlexNet $\rightarrow$ ResNet50, for which compute-augmenting algorithmic progress outstripped data-augmenting by 11-fold to AlexNet $\rightarrow$ VIT-e for which the same multiple was roughly 7x). This result is consistent with the idea that compute-augmenting algorithmic innovations have become relatively less important as compute budgets themselves have expanded sufficiently fast to offset the need for such innovations.

\subsection{Algorithmic progress doubles effective compute budgets roughly every nine months}
\label{sec:algo_progress}

Algorithmic advances are sometimes reported in terms of the rate at which less compute is needed to train a model to achieve the same level of performance (see notably \cite{hernandez2020measuring}). That is, if algorithmic innovations occur each year that enable us to train models that achieve some fixed level of performance with half the amount of compute used in the previous year, we might say `effective compute' doubles every 12 months.

Given that, effective compute $C$ is given simply as:
\begin{equation}
    C= \log\big(\exp(\alpha_1 + \alpha_{\textrm{Year}} \times (\textrm{Year} - 2012))\times \textrm{compute}^{\alpha_{\textrm{compute}}}\big)
\end{equation}
The doubling time of effective compute is simply the time it takes until algorithmic progress contributes $2^{-\alpha_{\textrm{compute}}}$ worth of effective compute, i.e., $\log(2) \times \alpha_{\textrm{compute}}/\alpha_{\textrm{Year}}$ years. By bootstrapping our estimates of these coefficients, we find a mean doubling time of 8.95 months [95\% CI: 3.55 months, 25.40 months]. In other words, algorithmic improvements amount to a doubling of effective compute roughly every 8.95 months.

Our estimates are consistent with, yet slightly shorter than, \cite{hernandez2020measuring}'s estimate that the compute required to achieve AlexNet's level of performance has halved every 16 months from 2012 to 2020. We expect that this difference exists for two key reasons that are worth elaborating on. Firstly, as we discuss in Section 2, their method is sensitive to choosing performance thresholds, and specifically, by choosing a relatively low performance threshold (36.7 top-1 accuracy, i.e., AlexNet's performance) their approach has the tendency to underestimate algorithmic progress relative to if they were to choose a higher performance threshold.

Secondly, there are additional compute-algorithmic improvements that our approach captures that \cite{hernandez2020measuring} might fail to capture, notably improvements in 'software-for-hardware' such as improvements in vendor libraries for GPU implementations like cuBLAS and cuDNN, which provide optimized implementations for standard routines such as forward and backward convolution, pooling, activations, and 
standard normalization techniques. Moreover, the introduction of novel high-level libraries such as DeepSpeed (\cite{rasley2020deepspeed}), interpreters and compilers such as Triton (\cite{tillet2019triton}) can provide substantial improvements in hardware performance. Hence, this suggests that re-implementations of the sort found in \cite{hernandez2020measuring} may unwittingly benefit from what might be substantial software improvements, such that the extent of software improvements for which their method controls might be a strict subset of all software improvements, therefore introducing a negative bias.

\section{Conclusion}
\label{sec:conclusion}

Measuring the extent of algorithmic progress in machine learning is tricky, as these typically resist more standard analyses grounded in algorithmic complexity theory. We developed an empirical approach—guided by empirical scaling laws—that enables us to directly estimate the effective compute and data budgets, and produce decompositions of the various drivers of progress computer vision. By estimating the marginal expected contributions from different sources of progress, we find, consistent with prior work on hardware and software in scientific computing, improvements in algorithms and hardware contribute roughly equally to performance improvements in computer vision. Our work indicates that, in particular, compute-augmenting algorithmic improvements (algorithmic innovations that enable researchers to do more with less compute), is the main driver of algorithmic improvements, at least for the computer vision models considered in this work. While past work has found that algorithmic innovation halves the compute requirements needed to train well-known computer vision models every 16 months, we have found that this is likely faster; we estimate this occurs every nine months. 

\section*{Appendices}
\label{sec:Appendix}

\subsection*{Appendix A: Estimating compute}
\label{sec:estimatingcompute}

We use two methods for inferring the amount of compute used to train an AI system, both are from \cite{sevillacompute}. These methods (one based on the architecture of the network and number of training batches processed, and another based on the hardware setup and amount of training time), are outlined below.

\subsubsection*{Method 1: Counting operations in the model}

The first method can be summarized as:
\begin{equation}
    \text{Training compute} = (\text{FLOP per forward pass} + \text{FLOP per backward pass}) \times \text{Nr. of passes},
\end{equation}
where $\text{FLOP per forward pass}$ is the number of floating point operations in a forward pass, $\text{FLOP per backward pass}$ is the number of operations in a backward pass and $\text{Nr. of passes}$ is the number of full passes (a full pass includes both the backward and forward pass) made during training.

$\text{Nr. of passes}$ is just equal to the product of the number of epochs and the number of examples: 
\begin{equation}
 \text{Nr. of passes} =\text{Nr. of epochs} \times \text{Nr. of examples}.
\end{equation}
If the number of examples is not directly stated, it sometimes can be computed as the number of batches per epoch times the size of each batch: $\text{Nr. of examples} = \text{Nr. of batches} \times \text{batch size}$  

Moreover, since computing the backward pass requires each layer to compute the gradient with respect to the weights and the error gradient of each neuron with respect to the layer input to backpropagate. Each of these operations requires compute roughly equal to the amount of operations in the forward pass of the layer. Hence, the ratio of the FLOP per forward pass to the FLOP per backward pass is 1:2. Thus we can simplify (14) as:
\begin{equation}
    \text{Training compute} = \text{FLOP per forward pass} \times 3 \times \text{Nr. of epochs} \times \text{Nr. of examples}.
\end{equation}

\subsubsection*{Method 2: GPU time}

Secondly, we may also use the reported training time and GPU model performance to estimate the training compute.

GPU-days describe the accumulated number of days a single GPU has been used for the training. For example, if the training lasted two days and a total of five GPUs were used, that equals 10 GPU-days. By multiplying the number of GPU-days by the performance of the GPU, we can infer the amount of compute, in FLOP, that was needed to train the model. In particular, the formula used is the following:
\begin{equation}
    \text{Training compute} = \text{Training time (in seconds)} \times \text{Nr. of cores} \times \text{Peak FLOP/s} \times \text{utilization rate},
\end{equation}
where peak performance in FLOP/s is found in the relevant GPU datasheets, and the utilization rate corrects for imperfect hardware utilization, for which 0.3 is often a reasonable baseline (see \cite{sevillacompute}), which is the utilization rate that we often assumed in case we could not otherwise derive it.

\subsection*{Appendix B: Empirical challenges}
\label{sec:challenges}

One of the main challenges we had while trying to fit the various models to the data is that due to the inadequate size and quality of the dataset, the log likelihood of the nonlinear models often have multiple local maxima that are far apart but give similar levels of goodness of fit to the data. In other words, the estimation suffers from a lack of concavity. As a result of this, standard errors obtained, for example, by inverting the Fisher information matrix are unreliable and inconsistent.

As such, we rely on bootstrapping to generate standard error estimates. The result is that depending on the exact model chosen, anywhere from \( 2 \% \) to \( 5 \% \) of the time one of the other local maxima of the log likelihood end up being preferred, which blows up the standard error estimates of the parameters. However, the parameter distributions obtained this way are not unimodal and are highly skewed, so the standard errors can offer poor guidance if they are interpreted as corresponding to, e.g., a normal distribution. To remedy this problem, we report \textit{both} standard errors \textit{and} \( 90 \% \) confidence intervals for all of the parameter estimates, though the nontrivial dependence structure between the parameters means the best way to take parameter uncertainty into account is again by bootstrapping.

The choice of a constant rate of improvement for compute and data use efficiency coming from algorithmic progress is made consciously with an associated bias-variance trade-off in mind. If we had a dataset that contained a sufficient number of images for each year, a method with less bias but more variance such as having a dummy variable for the algorithmic efficiency of each year would have been preferable. The limited size of our dataset and the heavy concentration of our data towards later years, however, means that such a method produces undesirably high variance in our estimates. Our conclusions should therefore not be interpreted as implying that the pace of algorithmic innovation has actually been constant: our data is simply not sufficiently high quality to answer such questions about higher-order trends in algorithmic progress with any degree of reliability.

\subsection*{Appendix C: Senstivity of threshold-based estimates}
\label{sec:Sensitivity}

In Section 3, we argued that a problem with \cite{hernandez2020measuring}'s definition of algorithmic progress is that it is highly sensitive to the exact benchmark and threshold pair that we choose. Here, we illustrate this with examples. 
\begin{table}[h!]
\centering
\begin{tabular}{@{}cc@{}}
\toprule
\begin{tabular}[c]{@{}c@{}}Performance threshold\\ (top-1 accuracy)\end{tabular} & \begin{tabular}[c]{@{}c@{}}Halving time\\  (months)\end{tabular} \\ \midrule
0.60 & 7.72 \\
0.70 & 7.46 \\
0.80 & 6.71 \\
0.84 & 5.35 \\
0.88 & 0.67 \\ \bottomrule
\end{tabular}
\caption*{\small \textbf{Table 4.} Estimated halving-time of the compute required to train models achieving a given level of performance.}
\end{table}
Hence, we show that the estimated halving time can differ by an order of magnitude, depending on whether an easy performance threshold, such as the performance of early models like \cite{krizhevsky2012imagenet}'s AlexNet (which achieves a top-1 accuracy of around 0.63), or the performance of recent top models, such as those of \cite{jiang2021all}, that achieve a performance of around 87\% without additional training data.

\newpage

\subsection*{Appendix D: Predictions}
\label{sec:Predictions}

We've only focused on the point estimates from the model so far. However, it's important to highlight that the model has high uncertainty in some situations, both due to the parameter uncertainty and due to the noise that's built into the model.\par

\begin{figure}[H]
\centering
\includegraphics[scale=0.3]{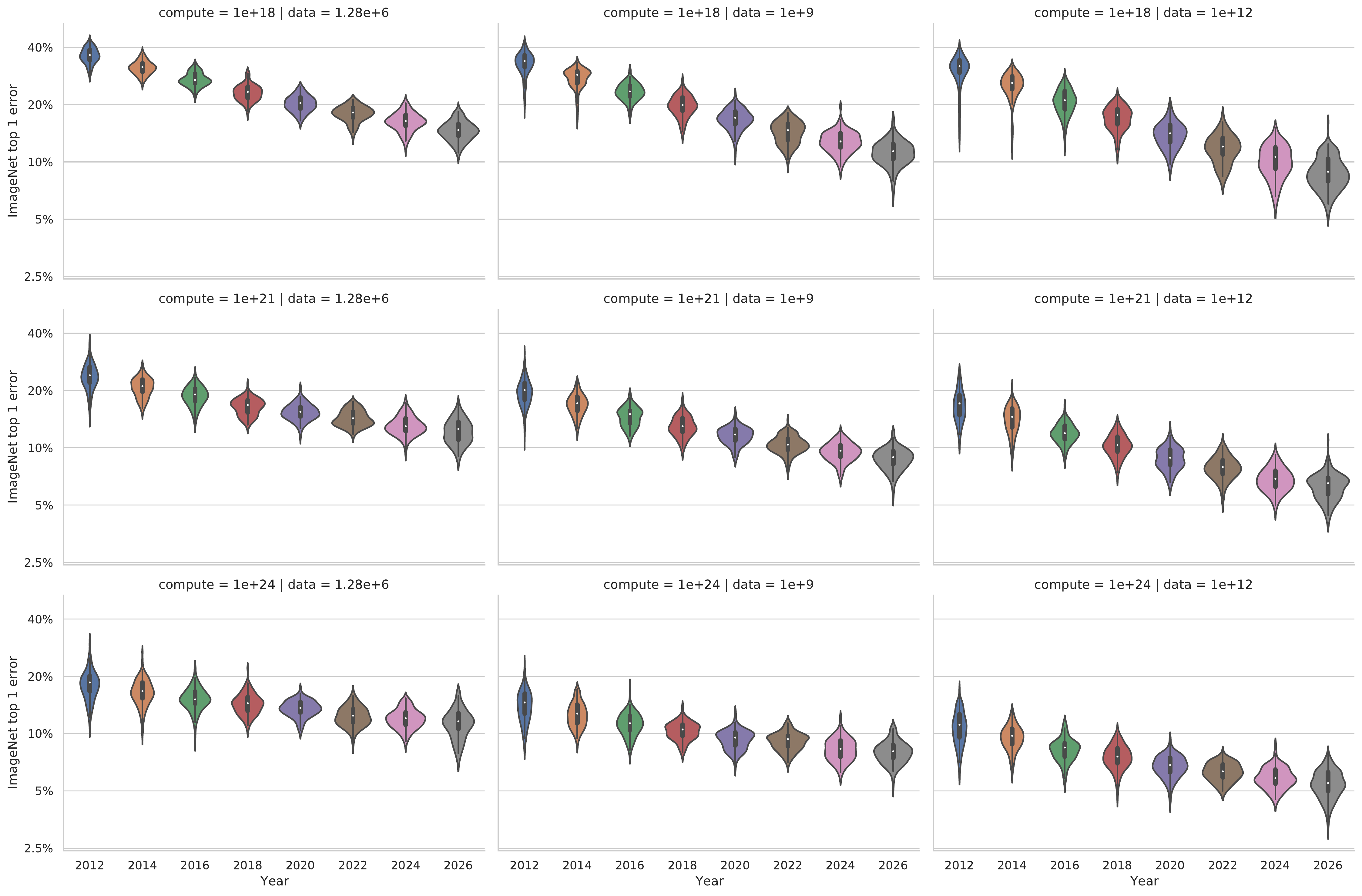}
\caption*{\textbf{Figure 3.} A grid of violin plots showing the model's predictions.}
\end{figure}

Figure 3 does a good job illustrating both the model's mean predictions and its uncertainty about its predictions. One notable pattern is the impact of time on models with an abundance of data versus an abundance of compute: the improvement is considerably faster in the compute bottlenecked model in the top right corner compared to the data bottlenecked model in the bottom left. In other words, while algorithmic improvements help with both compute and data bottlenecks, their impact on compute bottlenecks seem more significant than their impact on data bottlenecks.

\subsection*{Appendix E: Relationship with standard scaling laws}
\label{sec:scaling}

To illustrate how the model specifications we use relates to the usual form in which scaling laws are cast, recall that we can approximate the sigmoid function when the argument is large as

\[ \sigma(x) = \frac{k}{1 + e^{-x}} \approx k - e^{-x} \]

On our models, \( P \) is modelled as the product of two sigmoid terms. Hence, we get
\begin{align}
P &= \sigma(\beta_1 + \beta_{\text{year}}\times \text{year} + \beta_\text{data} \log(\textrm{Data})) \sigma((\alpha_1 + \alpha_{\textrm{year}} \times \textrm{year} + \alpha_\text{compute} \times \log(\textrm{compute}))) \\
&\approx (1 - e^{-(\beta_1 + \beta_{\text{year}}\times \text{year} + \beta_\text{data} \log(\textrm{Data}))})) (1 - e^{-(\alpha_1 + \alpha_{\textrm{year}} \times \textrm{year} + \alpha_\text{compute} \times \log(\textrm{compute}))}) \\
&= (1 - e^{-(\beta_1 + \beta_{\text{year}}\times \text{year})} D^{-\beta_{\textrm{data}}}) (1 - e^{-(\alpha_1 + \alpha_{\textrm{year}} \times \textrm{year}} C^{-\alpha_\textrm{compute}}) \\
&\approx 1 - (A^{\tilde{\beta}} D^{-\beta_\textrm{data}} + B^{\tilde{\alpha}} C^{-\alpha_\textrm{compute}}),
\end{align}
which makes the similarity to a traditional scaling law, of the type found in \cite{hoffmann2022training}, apparent.

\subsection*{Appendix F: Supporting results}
\label{sec:supporting}

\subsubsection*{Decompositions in log-odds space}
In this section, we re-produce our earlier decompositions, but instead of considering linear distance in predicted performance, i.e., $|f(\textrm{Year}_1, \textrm{Data}_1, \textrm{Compute}_1)- f(\textrm{Year}_2, \textrm{Data}_2, \textrm{Compute}_2)|$, we now consider distance in `log-odds space', i.e., $|\text{logit}(f\big(\textrm{Year}_1, \textrm{Data}_1, \textrm{Compute}_1)\big) - \text{logit}\big(f(\textrm{Year}_2, \textrm{Data}_2, \textrm{Compute}_2)\big)|$, which is a natural distance-measure—often used in probability-theory—for numbers bounded on the unit interval.

\begin{table}[h!]
    \centering
\begin{tabular}{@{}lcccc@{}}
\toprule
 & \multicolumn{1}{l}{\begin{tabular}[c]{@{}l@{}}Reduction\\ in error\end{tabular}} & \multicolumn{1}{l}{\begin{tabular}[c]{@{}l@{}}Algorithmic \\ progress\end{tabular}} & \multicolumn{1}{l}{\begin{tabular}[c]{@{}l@{}}Compute \\ scaling\end{tabular}} & \multicolumn{1}{l}{\begin{tabular}[c]{@{}l@{}}Data \\ scaling\end{tabular}} \\ \midrule
AlexNet $\rightarrow$ ResNet50 & 23.7 & 65.0\% & 35.1\% & NS \\
AlexNet $\rightarrow$ ResNeXt-101 & 24.2 & 70.6\% & 29.2\% & NS \\
AlexNet $\rightarrow$ BiT-L & 24.2 & 39.9\% & 46.6\% & 12.1\% \\
AlexNet $\rightarrow$ VIT-H/14 & 24.8 & 43.0\% & 46.6\% & 13.5\% \\
AlexNet $\rightarrow$ VIT-e & 27.6  & 40.2\% & 42.6\% & 17.2\% \\
ResNet50 $\rightarrow$ BiT-L & 10.4 & 29.9\% & 47.2\% & 22.9\% \\
ResNet50 $\rightarrow$ VIT-H/14 & 10.9 & 34.4\% & 43.2\% & 22.4\% \\
ResNet50 $\rightarrow$ VIT-e & 13.8 & 32.7\% & 40.6\% & 26.7\% \\
ResNeXt-101 $\rightarrow$ BiT-L & 6.6 & 24.0\% & 49.8\% & 26.2\%\\
ResNeXt-101 $\rightarrow$ VIT-H/14 & 7.2 & 29.3\% & 45.3\% & 25.4\% \\ 
ResNeXt-101 $\rightarrow$ VIT-e & 10.0 & 28.9\% & 41.5\% & 29.6\% \\ \bottomrule
\end{tabular}
    \caption*{\centering \small \textbf{Table 2.} Attribution of progress to algorithmic progress, compute scaling and data scaling between model pairs based on Shapley decomposition in log-odds space. ``NS" indicates that there was no scaling of the relevant input between these models. Numbers may not all add up to 100 due to rounding.}
    \label{tab:my_label}
\end{table}

\begin{table}[h!]
    \centering
\begin{tabular}{@{}lcccc@{}}
\toprule
 & \multicolumn{1}{l}{\begin{tabular}[c]{@{}l@{}}Reduction\\ in error\end{tabular}} & \multicolumn{1}{l}{\begin{tabular}[c]{@{}l@{}}Algorithmic progress\\(data augmenting)\end{tabular}} & \multicolumn{1}{l}{\begin{tabular}[c]{@{}l@{}}Algorithmic progress\\(compute augmenting)\end{tabular}} & \multicolumn{1}{l}{\begin{tabular}[c]{@{}l@{}}\end{tabular}} \\ \midrule
AlexNet $\rightarrow$ ResNet50 & 23.7 & 5.3\% & 59.7\%\\
AlexNet $\rightarrow$ ResNeXt-101 & 24.0 & 6.1\% & 64.7\% \\
AlexNet $\rightarrow$ BiT-L & 24.2 & 5.5\% & 34.4\% \\
AlexNet $\rightarrow$ VIT-H/14 & 24.8 & 6.1\% & 36.9\%  \\
AlexNet $\rightarrow$ VIT-e & 27.6  & 6.5\% & 33.7\% \\
ResNet50 $\rightarrow$ BiT-L & 10.4 & 5.3\% & 24.6\% \\
ResNet50 $\rightarrow$ VIT-H/14 & 10.9 & 6.3\% & 28.1\% \\
ResNet50 $\rightarrow$ VIT-e & 13.8 & 6.8\% & 25.9\% \\
ResNeXt-101 $\rightarrow$ BiT-L & 6.6 & 4.5\% & 19.5\% \\
ResNeXt-101 $\rightarrow$ VIT-H/14 & 7.2 & 5.7\% & 23.6\% \\ 
ResNeXt-101 $\rightarrow$ VIT-e & 10.0 & 6.4\% & 22.5\%  \\ \bottomrule
\end{tabular}
    \caption*{\centering \small \textbf{Table 3.} Shares of algorithmic progress that is compute- vs. data-augmenting when decomposition is performed in log-odds space.}
    \label{tab:my_label}
\end{table}

It is notable that these results do not substantially differ from our earlier results, i.e., when the decomposition was done by applying Shapley values in linear space. This indicates that our key results are robust to this change in the analysis.

\newpage

\printbibliography

\end{document}